# Multi-Phase Multi-Objective Dexterous Manipulation with Adaptive Hierarchical Curriculum

Lingfeng Tao*, Jiucai Zhang^, Xiaoli Zhang*, *Member, IEEE*

*Abstract*— Dexterous manipulation tasks usually have multiple objectives, and the priorities of these objectives may vary at different phases of a manipulation task. Varying priority makes a robot hardly or even failed to learn an optimal policy with a deep reinforcement learning (DRL) method. To solve this problem, we develop a novel Adaptive Hierarchical Reward Mechanism (AHRM) to guide the DRL agent to learn manipulation tasks with multiple prioritized objectives. The AHRM can determine the objective priorities during the learning process and update the reward hierarchy to adapt to the changing objective priorities at different phases. The proposed method is validated in a multi-objective manipulation task with a JACO robot arm in which the robot needs to manipulate a target with obstacles surrounded. The simulation and physical experiment results show that the proposed method improved robot learning in task performance and learning efficiency.

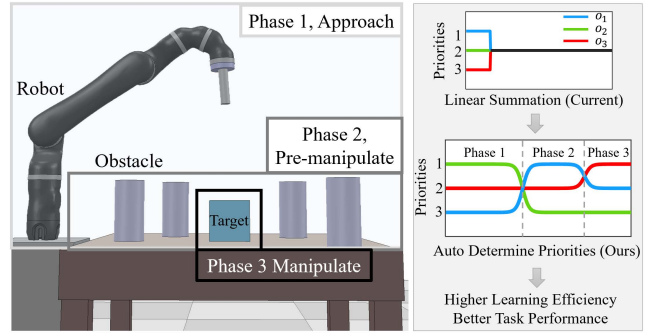

Fig. 1. A manipulation task with three phases: (1) approach; (2) pre-manipulate; (3) manipulate. Three task objectives are defined: avoid obstacles ($o_1$); minimize the execution time ($o_2$); manipulate ($o_3$). The objective priorities are expected to change in different phases. While a current reward mechanism usually uses linear summation, which is inefficient to learn and hard to find a good policy. In addition, the robot should determine the objective priorities during the learning process to improve its learning efficiency and policy performance.

## I. INTRODUCTION

Dexterous manipulation is essential to increase robots' usability in tasks such as assembly, healthcare, education, and living assistance. These tasks typically need to be finished in multiple phases, and each phase has multiple objectives. Although all phases usually share the same set of objectives, the priorities of objectives in each phase can vary, which are critical to achieving the manipulation tasks' efficiency and success rate. For example, an assembly task usually has two phases: (1) approaching, (2) installation. All phases share three objectives: (a) fast assemble speed, (b) high assembly quality, and (c) avoid the collision. In the first phase, the robot picks up the assembly part and move to the target position. The task objective with the top priority is to avoid touching other parts, then try to move faster to minimize the execution time, and the lowest priority is to improve the installation quality. In the second phase, the robot reaches the target position and is ready for installation, and now the top priority changes to improve the installation quality, then minimize the execution time and avoid touching other parts.

Existing research in the traditional control method mainly focuses on how to weigh multiple objectives to balance objectives with optimization methods, which is computationally inefficient. Although reinforcement learning (RL) has been proven effective in enabling the robot to conduct autonomous manipulation tasks intelligently [1-5], the current reward formulation is usually a linear summation of the reward components of objectives, which is implicit and inefficient to learn the objective priorities, and causing poor learning performance (i.e., take a long time to learn or even fail to learn a correct policy). Furthermore, the current reward mechanism is usually fixed through all phases. This one-fix-all solution (i.e., using the same objective priority for all phases) cannot ensure each phase's local performance to be optimal. Such solutions may lead to suboptimal performance as the reward is not customized for each phase of the task. As a result, the learning performance and learning efficiency are usually limited in manipulation tasks where multiple objectives exist.

Consideration of objective priorities and priority changes across phases is essential to help the robot achieve optimal performance and speed up the learning efficiency. To achieve this goal, we develop the AHRM method to enable the robot to determine the objective priorities and adapt to the priority changes during different task phases. Our method's novelty is that the robot with AHRM can utilize the experience when exploring the environment to determine the objective priorities and update the reward hierarchy mechanism during the learning process. The benefits include better learning efficiency, higher task performance, and more appropriate policy behavior when comparing to the aforementioned existing RL methods. In summary, the main contributions of this work are as follow:
1) Solve the manipulation task in a DRL manner with a hierarchical reward mechanism to enable the robot to

*L. Tao and X. Zhang are with Colorado School of Mines, Intelligent Robotics and Systems Lab, 1500 Illinois St, Golden, CO 80401 USA (Phone: 303-384-2343; e-mail: tao@mines.edu, xlzhang@mines.edu).

^J. Zhang is with the GAC R&D Center Silicon Valley, Sunnyvale, CA 94085 USA (e-mail: zhangjiucai@gmail.com).

learn multiple objectives with different priorities efficiently.
2) Propose AHRM to consider both the objective priory changes and corresponding reward mechanism hierarchy to enable the robot to determine the objective priorities and adapt to the priority change across phases.
3) Validate the AHRM method and compare different objective priority determination methods (i.e., autonomously determined dynamic objective priorities, empirically defined dynamic objective priorities, fixed objective priorities, and linear summation reward) in a manipulation task.

## II. RELATED WORK

Traditional non-learning control methods have tried to solve the manipulation task when multiple objectives need to be done. With a stochastic optimization procedure [6], the priorities of each objective can be automatically determined as the task changes. The priorities are represented as parameterized weight functions to adjust the linear summation of the output (e.g., force or torque) [7]. These methods develop multiple controllers for multiple objective functions and adjust controller outputs' contributions to take objective priorities into account. In contrast, our method allows the robot to use a single policy to embed the knowledge of objective priorities into the control strategies enabled by the DRL, which reduces the robot's computational burden once the policy is learned.

DRL has been proven promising to enable autonomous manipulation, but the learning efficiency has been a great challenge. In [8], a DRL method combined off-policy updates and parallel training to reduce the training time and enable a physical robot to complete a door opening task. The work in [9] utilized the human demonstration to initialize the DRL policy to effectively reduce the training time to control the robot hand in a hammer usage task and an object relocation task. For a multi-objective manipulation task, [10] studied the compositionality of soft Q-learning methods in a multi-policy and multi-Q-function setup, which provides a method to construct new policies composing learned skills to improve learning efficiency. Despite the progress of existing research, learning efficiency in multi-objective dexterous manipulation tasks is still an open problem due to the difficulty of providing sufficient information for the reward design.

Although hierarchical reward methods exist that enable a robot to learn multi-objective tasks, these methods used a fixed objective priority for a given task, so does the reward hierarchy. The formulation of the reward hierarchies contains logical or weighted connections. Logical connections are strict constraints, where the higher-level hierarchy must be learned before the lower-level hierarchy. Weighted connections are soft constraints, where the higher-level hierarchy and lower-level hierarchy are learned together with a weighted summation. In [11], a genetic RL agent for swarm robot control is trained with a logically connected hierarchical reward function using a curriculum design method. A reward machine is used in [12] to learn the task structure and decomposition to the RL learner and achieves better learning efficiency than conventional Hierarchical RL methods.

Existing studies rarely consider updating the reward mechanism according to the objective priority changes during the different phases of the task. A similar approach is probably the reward shaping method that changes the reward function and provides the robot with an extra reward during the learning process to redirect the training [13]. Earlier approaches like [14] consistently provide an additional reward to the robot based on the state's potential. Later approaches start to discuss adaptive reward functions during the learning process. In [15], verbal feedback is used as an additional reward and is occasionally provided to the robot when it should adapt to new conditions. A dynamic reward function with adjustable parameters is proposed in [16] to adjust the reward function based on experience. However, current dynamic reward functions typically use a linear summed reward, which does not contain a hierarchical structure to guide the robot to determine objective priorities.

## III. METHODOLOGY

This section explains the development of AHRM (Algorithm 1) in a multi-objective, multi-phase task. The proposed method advances the reward function's design, which updates the reward structure during the learning process based on the objective priorities in each phase to enable the robot to understand and adapt to the objective priority changes in different phases of the task.

### A. Model Structure

We model the manipulation task as an RL problem that follows the MDP. The MDP is defined as a tuple $\{S, A, R, \gamma\}$, where $S$ is the state of the environment, $A$ is the set of actions. $R(s'|s, a)$ is the reward function to give the reward after the transition from state $s$ to state $s'$ with action $a$. $\gamma$ is a discount factor. A policy $\pi(s, \theta)$ specifies the action for state $s$. $\theta$ is the policy network parameters. A Proximal Policy Optimization (PPO) algorithm [17] is adopted to find $\theta$.

The reward function $R$ is the design target of AHRM. R is the key component to guide the robot to learn the prioritized objectives in each phase in a multi-objective and multi-phase task. We assume that the task objectives and the task phases can be heuristically defined by the developers based on empirical knowledge and human perception. However, it is challenging for human experts to define appropriate objective priorities for each phase due to individual experience and bias. Thus, the robot should determine the priorities of objectives based on its own experience. For each objective, a reward component $f_i$ is defined, where $i=1, 2, ..., N$. $N$ is the total number of the task's objectives. Each objective can only have one priority level, which is denoted as $j$, where $j=1, 2, ..., N$. The priority level follows a descending manner as $j$ increases (i.e., the objective with priority 1 has the highest level). The phases of the task are denoted as $k=1, 2, ..., K$. $K$ is the number of phases. Each phase's reward function is constructed with all reward components in a hierarchical mechanism, which is introduced in the following section.

### B. Hierarchical Reward Mechanism for Each Phase

The learning of the hierarchy in a phase $k$ follows the same descending order as the objective priorities, which means the objectives with higher priorities are first learned. We define the hierarchical reward mechanism $\mathbb{R}^k = [R_1^k, ..., R_j^k, ..., R_N^k]$. $R_j^k$ is the reward function for hierarchy level $j$ (the same with objective priorities), which is computed as:

$$R_j^k = \sum_{m=1}^{j} f_i|m, \quad j=1,2,...,N \quad (1)$$

where $f_i|_m$ is the reward component for objective $i$ with priority $m$. When the robot finishes learning a hierarchy level $j$ with reward $R_j^k$, it will learn the next hierarchy level by moving to the next reward component $R_{j+1}^k$. The criterion of successfully learning the hierarchy level $j$ in phase $k$ is to check if $\sum R_j^k$ has converged.

The learning sequence of hierarchical reward mechanism $\mathbb{R}^k$ is not strong enough to avoid interference with the higher-level priorities while learning the lower-level priorities (e.g., alter the previously learned policy). To avoid this interference, the learning process is subject to a hard priority constraint. The criterion is that the reward components of learned higher-level objectives become constraints for the following episodes to avoid interference between hierarchy levels. When learning an objective with hierarchy level $j$, we define the constraints as:

$$\sum R_{j-1}^k \geq \tau_{j-1} \quad (2)$$

where threshold $\tau_{j-1}$ is the average of the episode reward $\sum R_{j-1}^k$ in the last 10 episodes. When Eq. 2 is violated, the current episode is terminated, and a new episode starts.

### C. Determination of Objective Priorities in Each Phase

The goal of this section is to find the optimal hierarchy for each phase. That is, to select the objective for each priority. In our work, each training episode is expected to explore all the phases of a given task and find the priorities and their changes across different phases. However, at the start of the DRL training, a single episode may not complete all the phases. That is, one episode may fail in an intermediate phase of the task. To ensure the robot has sufficient information to determine objective priorities for a specific phase, the hierarchy can be determined and updated only when that phase is explored for $P$ times, and $P$ is the empirically defined threshold. Specifically, we count the number of times $e_k$ that the robot visits a phase $k$, If $e_k \geq P$, the robot will determine the objective priorities of phase $k$ based on the collected information.

There can be multiple strategies to determine the objective priorities, such as the objectives with the least collected reward should be learned first, or the objectives with the highest collected reward should be learned first. In this work, we define the objective with a higher absolute value of the average episode reward should be learned first because the robot receives more information for that objective no matter it is positive or negative. We calculate the average episode reward of each objective:

$$\bar{r}_i^k = \frac{1}{P} \sum_{p=1}^{P} r_{i,p}^k \quad (3)$$

where $r_{i,p}^k = \sum f_i|k,p$ is the episode reward for objective $i$ in phase $k$ at episode $p$.

The priorities for the objectives are updated by sorting the average cumulative rewards. To ensure the sorting operation's veracity, the magnitude of the reward components should be consistent along the time horizon, which means that the reward components need to be normalized or be defined with the same principles. Then, we extract the index of the sorted absolute average episode reward for each objective:

$$[I_1^k, ...I_j^k..., I_N^k] = Index\left[\underset{high \to low}{sort} |\bar{r}_i^k|\right] \quad j=1,...,N \quad (4)$$

where $I_j^k$ is the index of the $\bar{r}_{I_j^k}^k$. Finally, the reward component of the priority $j$ is assigned as $f_{I_j}$. That is

$$f_i\Big|j = f_{I_j}\Big|j \quad (5)$$

We assume that there is only one optimal hierarchy for each phase. Once priorities of all objectives are determined, the optimal hierarchy is found for this phase. The robot will use this hierarchy in the rest of the learning process.

### D. Smooth Transition of Reward Hierarchy across Phases

When the robot needs to adapt to the dynamic objective priorities in different phases, the current phase's reward hierarchy transits to a different one in the next phase. To mitigate the negative impact to the policy performance when reward changes, a smooth transition function from one hierarchy to another is designed by continuously updating the hierarchical reward function. For easy modification, monotonically bounded functions are better choices. In this work, we choose a *tanh* function as the smooth transition function:

$$tanh(x) = \frac{e^{2x}-1}{e^{2x}+1} \quad (6)$$

where $x$ is the state input. Then the reward function becomes:

$$R_j^k = \begin{cases} \sum_{m=1}^{j} f_{i'}|m[\frac{1}{2}tanh(\alpha x)] + \sum_{m=1}^{j} f_i|m[-\frac{1}{2}tanh(\alpha x)] & x \leq \delta \\ \sum_{m=1}^{j} f_{i'}|m & x > \delta \end{cases} \quad (7)$$

where $i'$ is the updated objective index with priority level $j$ after the phase change, $i$ is the previous objective index with priority $j$ before the phase change. $\alpha$ is the adjustable parameter to control the slope and range of the smooth function. $x$ is calculated as:

$$x = \sqrt{\sum [s_t - b(k,k+1)]^2} \quad (8)$$

$x$ the Euclidean distance between the current state $s_t$ and the predefined phase boundary $b(k,k+1)$. $\delta$ is the threshold of the transition period between phases, which can be empirically defined.

---

**Algorithm 1** AHRM

1: **procedure** Initialize agent policy $\pi_0 = \pi(\theta_0)$, read reward components $f_i$, $i = 1, 2, ..., N$. Initialize objective priorities $j_k \ \forall k$.
2:    **for** *episode=1,2,...* **do**
3:       obtain initial state $s_1$
4:       if $e_k > E$, update objective priorities $j_k$ with Eq. 4 and Eq. 5
5:       read objective priorities $j_k \ \forall k$, build reward function and its transition with eq. 6
6:       **for** $t=1,2,...,T$ **do**
7:          Select action $a_t$ with policy $\pi(s_t, \theta_{t-1})$
8:          **return** trajectory $[s_1, a_1, s_2, a_2, ..., s_T, a_T]$
9:       **end for**
10:     if constraints in eq. 2 are satisfied, update policy $\pi(\theta_t)$, else, restart episode
11:     count $e_k$ for phase $k$
12:    **end for**
13: **end procedure**

## IV. EXPERIMENTS

The experiment was implemented in a simulated environment using V-REP [18] and a physical environment. Specifically, the task environment contains a JACO arm [19] with a cylindroid end effector attached to the end joint (Fig. 2, in the physical environment, the JACO arm just keeps its hand close for simplicity). The JACO arm is placed next to a table. The target is a cube that is on the top of the table and located at the center. The other objects on the table are obstacles. The robot's task is to use the end effector to push the target off the table without touching the obstacles. Three objectives are defined to be satisfied by the robot for optimal performance. To ensure the consistency of sorting operation in eq. 4, the corresponding reward functions are defined in binary form:

(1) Avoid obstacles:

$$f_1 = \begin{cases} -1 \text{ if } \forall V_{obstacle} > 0 \\ 1 \text{ if } \forall V_{obstacle} = 0 \end{cases} \quad (9)$$

where $V_{obstacle}$ is the speed of obstacles. The function returns a penalty if the robot or target touches any of the obstacles. Otherwise, it returns 0.

(2) Manipulation:

$$f_2 = \begin{cases} -1 \text{ if } d-l \leq d'-l' \\ 1 \text{ if } d-l > d'-l' \end{cases} \quad (10)$$

where $d$ is the current distance of the cube from the table center, $d'$ is the same property in the last time step. It returns a positive reward to encourage the robot to push the target away from the table center until it falls. $l$ is the current distance between the end effector and the cube center, $l'$ is the same property in the last time step.

(3) Minimize the execution time:

$$f_3 = \begin{cases} -1 \text{ if } t_k < \Gamma_k \\ 1 \text{ if } t_k = \Gamma_k \end{cases} \quad (11)$$

where $t_k$ is the time spent in phase $k$, $\Gamma_k$ is the episode length in the single-phase $k$ at the last episode. The term is a penalty component based on time to encourage the robot to act quickly towards the target. The objective priorities are determined every 30 episodes.

The action space includes the 5 joints of the JACO arm. The state observation includes the joints angle, target position, target velocity, obstacle position, obstacle velocity, and previous action. Each episode was set to 60 seconds to offer the robot enough time to explore the environment and finish the task. The deep neural network of the PPO agent has three fully connected layers, and each layer has 1000 neurons. The hyper-parameters of PPO algorithms are presented in Table I.

TABLE I. HYPER-PARAMETERS FOR PPO ALGORITHM

| Parameters | Values |
|---|---|
| Discount Factor($\gamma$) | 0.995 |
| Experience Horizon | 512 |
| Entropy Loss Weight | 0.02 |
| Clip Factor | 64 0.05 |
| GAE Factor | 0.95 |
| Sample Time | 0.05 |
| Mini-Batch Size | 64 |
| Learning Rate | 0.001 |
| Number of Epoch | 3 |

Based on the task environment, three phases are defined in spatial order (Fig. 1), and each phase covers a portion of the 3D task space. The phase identification is based on the position of the end effector. The first phase starts from the initial position and covers all the space outside of the obstacles' boundary. The second phase is the space in the boundary of the obstacles before touching the target cube. The third phase is a cubic space around the target.

To evaluate the AHRM method's performance, two other DRL methods are designed with the hierarchical reward mechanism. The first DRL method uses a hierarchical reward mechanism with empirically defined task objectives in different phases (denoted as MHRM), shown in Fig. 3(b). The second DRL method uses a hierarchical reward with pre-defined fixed objective priorities to train the agent (denoted as FHRM), shown in Fig.3(c). The baseline method trains a DRL agent with a reward function using an equally weighted linear summation (denoted as LS), shown in Fig. 3(d).

## V. RESULTS AND DISCUSSION

Figure 3(a) shows the learned objective priorities with AHRM for each phase. The training progress of all the methods is shown in Fig. 4 and in the accompanying video. All methods are trained for 300 episodes. The visualizations of the learned policies are shown in Fig. 5-Fig, 7. Each policy presents 6 screenshots show the critical moments in the task.

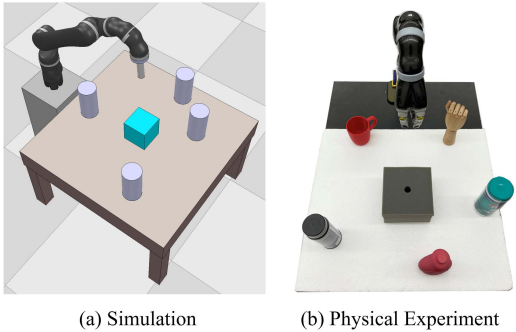

(a) Simulation    (b) Physical Experiment

Fig. 2. Experiment setup. A JACO robot arm is attached with a cylindroid end effector next to a table. The target is a cube that is located at the center top of the table, the others as obstacles. The task is to push the target cube off the table and avoid touching the obstacles.

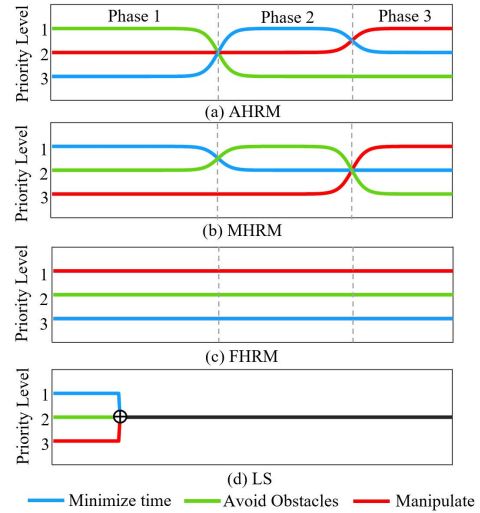

Fig. 3. (a) AHRM: the objective priorities for each phase are determined by the robot. (b) MHRM: the tasks priorities and their change for each phase are empirically defined. (c) FHRM: the objective priorities are empirically defined and fixed. (d) LS: the reward is linearly summed.

Table II shows the performance statistics of the trained policies. The training progress shown in Fig. 4 shows that the AHRM method outperformed the other three methods in learning efficiency. Comparing to MHRM, AHRM determined different objective priorities in phase 1 and phase 2, shown in Fig. 3. In the experiment, the priority determination strategy is to learn the objective with a higher average reward. In phase 1, the robot can never touch an obstacle, so the reward received for this objective is always the highest. Thus, avoid obstacles became the top priority. For phase 2, minimize time became the top priority. One possible explanation is that the robot spent more time adjusting its pose

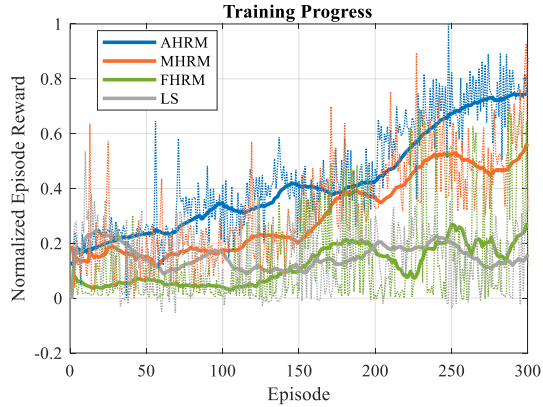

Fig. 4. Comparison of the learning progress for all the tested methods. All methods are trained 300 for episodes.

TABLE II. PERFORMANCE STATISTICS OVER 40 EVALUATIONS

| Method | S* | S Rate | Touch Obstacles | Time (s) | Cube Travel Length (m) |
|---|---|---|---|---|---|
| AHRM | 37 | 92.5% | 2 | 3.8±0.3 | 0.5103±0.0032 |
| MHRM | 34 | 85% | 5 | 4.4±0.6 | 0.5148±0.0024 |
| FHRM | 30 | 75% | 36 | 6.5±1.3 | 0.5274±0.0153 |
| LS | 4 | 10% | 39 | 8.3±1.1 | 0.5342±0.0219 |

* *S* is success. *S Rate* is the success rate among 40 evaluations. *Touch Obstacles* counts the total times that the robot touches the obstacles (rounded). *Time* is the average time spent until task success. *Cube Travel Length* is the average trajectory length of the cube.

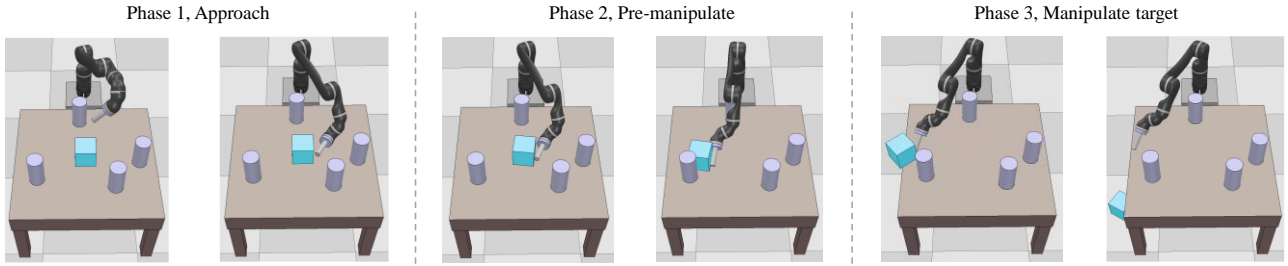

Fig. 5. Visualization of the AHRM policy that has learned to push the target cube down the table and avoid touching the obstacles. The robot learned to approach the target from the right side, adjust the end-effector pose to prepare for manipulation, and then push the target down the table. The policy learned with the MHRM method has similar behaviors but with a longer training time and a slightly worse task performance, as shown in Table I. For simplicity, the MHRM visualization is not shown.

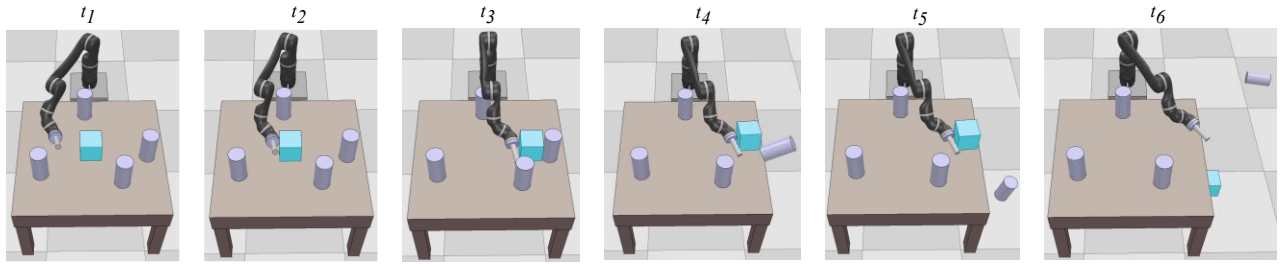

Fig. 6. Visualization of the FHRM policy. The robot approached the target from the left side, then adjusted the pose and pushed down the target. Its behavior (i.e., approaching the target from the left side) is different from AHRM and MHRM, which has a higher chance to touch an obstacle and knocked it off.

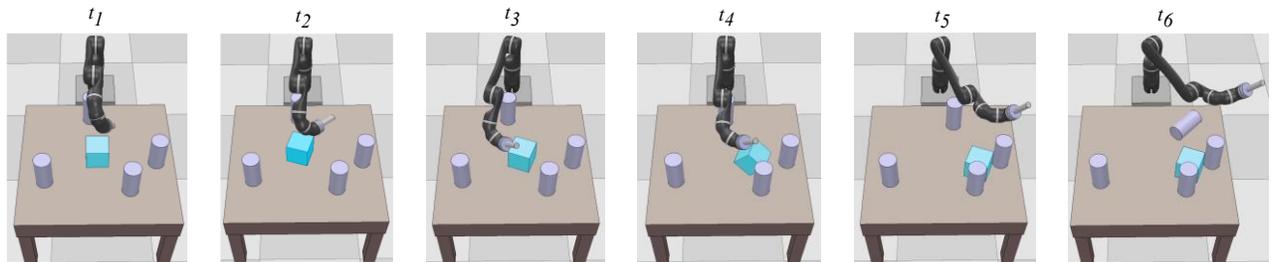

Fig. 7. Visualization of the LS policy. The robot approached the target from the top side. The end effector touched the target while it was adjusting pose, then caused the target to rotate before manipulating. As a result, the manipulation difficulty increased, and the robot failed the task.

to prepare for manipulation, which brings a large penalty in execution time. Thus, the robot tried to transit to phase 3 at a faster speed to manipulate the target. The determined objective priorities are different from MHRM. Our interpretation is that the priorities determined by AHRM are optimal for learning the task, while the empirically defined objective priorities in MHRM match human developers' performance expectations, but it is sub-optimal for the robot to learn. FHRM had difficulty maintaining the performance as the fixed objective priorities cannot achieve local optimal for each phase. The robot with the LS method struggled to find the policy due to the implicit priority information. The results of AHRM and MHRM in Table II prove that dynamically updating the objective priorities can effectively improve the learning efficiency and learning outcomes.

The policy evaluation shows that different reward mechanisms can result in different robot behaviors. For the designed task, it is not easy to manipulate the target in any direction due to the end effector's cylindrical shape. Any inappropriate contact may move the target to an uncomfortable position, like rotate the target or push it away. The robot with the AHRM learned that it should approach the target's side to align the end effector horizontally to maximize the contact surface and maintain stable contact points, as shown in Fig. 5. The robot also learned that approaching the right side of the target may achieve better performance even though the left side seems easier because there are fewer obstacles than the other side. Furthermore, the robot knows to slow down the action and make fine motion adjustments to get in contact with the target. Finally, the robot manipulates the target faster as it already learned that push to the left has less chance to touch obstacles. The robot with the MHRM method learned similar behaviors because the empirically defined objective priorities are near-optimal. The robot with FHRM learned different behaviors, which is to push the target to the right, as shown in Fig. 6. It can complete the task but has a higher chance to touch the obstacles. It is due to the fixed hierarchy cannot appropriately describe the objective priorities for all phases. As a result, the robot may only learn the top priority of the fixed hierarchy to push off the target but did not thoroughly learn the other two objectives. The LS method shown in Fig. 7 puts the robot's burden on finding the objective priorities during the learning process because the linear summed reward is abstract and implicit, which does not provide the information of objective priorities. As a result, the robot failed to find a stable policy in the learning process.

## VI. CONCLUSION

In this work, we propose an Adaptive Hierarchical Reward Mechanism method to help the robot determine and adapt to the objective priority changes while learning a multi-objective manipulation task. Overall, the training progress and the evaluation of the policy performance show that the AHRM method improves learning efficiency and task performance. Our future work will focus on studying the principles of how AHRM improves performance and optimizing the algorithm with different hierarchy structures and smooth transition functions.

ACKNOWLEDGMENT

This material is based on work supported by the US NSF under grant 1652454. Any opinions, findings, conclusions, or recommendations expressed in this material are those of the authors and do not necessarily reflect those of the National Science Foundation.